\documentclass{article}
\usepackage{multirow}
\usepackage{makecell}
\usepackage[preprint]{corl_2026} % Uncomment for pre-prints (e.g., arxiv); This is like ``final'', but will remove the CORL footnote.
\usepackage{graphicx}
\usepackage{tabularx}
\usepackage{booktabs}
\usepackage{caption}
\usepackage{array}
\usepackage{wrapfig}
\usepackage{needspace}
\usepackage{appendix}
\usepackage{amssymb}
\usepackage{amsmath}
\usepackage{algorithm}
\usepackage{bm}
\usepackage[noend]{algpseudocode}
\usepackage{arydshln}
\usepackage{hyperref}

\newcommand{\rev}{\textcolor{black}}

\title{PLUME: Probabilistic Latent Unified World Modeling and Parameter Estimation for Multi-Finger Manipulation}

\author{
  Abhinav Kumar$^{1}$ \And Soshi Iba$^{2}$ \And  Rana Soltani Zarrin$^{2}$ \And Dmitry Berenson$^{1}$\\
   $^1$University of Michigan  $^2$Honda Research Institute USA
}

\begin{document}
\newcommand{\namenospace}{\rev{PLUME}}
\newcommand{\name}{\namenospace~}
\maketitle

%===============================================================================
\begin{abstract}
    Dexterous manipulation with multi-finger hands can be sensitive to physical parameters such as object shape, pose, and friction coefficients.
    While simulation enables large-scale data collection with known parameter values, simulation-trained policies must still handle uncertainty at deployment, where the true parameters and therefore the true dynamics are unknown.
    \rev{Standard domain randomization strategies may be insufficient for precise tasks like screwdriver turning, as manipulation strategies may need to change depending on specific parameter values.}
    To address this, we propose Probabilistic Latent Unified world Modeling and parameter Estimation (\namenospace), a world model that jointly learns to evolve a belief over parameter values as well as \rev{the system dynamics conditioned on those parameters}.
    We learn a latent space to jointly represent multiple qualitatively different physical parameters along with rewards, themselves functions of \rev{partially-observable} variables, to inform planning.
    Our novel learning framework leads to efficient alignment of the world model to true dynamics through \rev{online} parameter inference as opposed to re-training or fine-tuning.
    We evaluate our method on simulated screwdriver turning, valve turning, bucket lifting, and disk flicking tasks, as well as a hardware screwdriver turning task, where we achieve successful zero-shot transfer of our simulation-trained policy and outperform state-of-the-art offline reinforcement learning and world-model-augmented behavior cloning baselines.
    Please see our website at \url{https://plume-world-model.github.io} for videos. 

\end{abstract}

\keywords{Robot manipulation, Probabilistic learning and uncertainty in robotics, dexterous manipulation with multi-finger hands} 

%===============================================================================

\vspace{-.3cm}
\section{Introduction}
\vspace{-.3cm}
\label{sec:intro}
	   As data-driven methods for performing fine manipulation tasks with multi-finger hands \cite{qi2023hand, kumar2025diffusing, kumar2025diffusion, wang2024penspin, liang2025dexhanddiff, yamadad, xu2025dexumi, ye2025dex1b, jiang2025dexmimicgen} improve, reliability and generalizability challenges remain.
    A core issue is the sensitivity of some dexterous manipulation tasks, such as screwdriver turning, to physical parameters such as friction coefficients or object shapes, which may be uncertain at deployment.
    \rev{For these precise tasks, manipulation strategies may need to change depending on the parameters, and thus domain randomization techniques designed to learn parameter-agnostic policies may be insufficient.
    }

    While recent world modeling methods \cite{zhu2025unified, ye2026world, maes2026leworldmodel, kim2026cosmos} offer great promise in using dynamics learning as a way to augment policy learning or to perform planning, these methods generally rely on a fixed-horizon history to implicitly account for dynamics variation and are generally not deployed to the types of precise multi-finger manipulation tasks we consider, which are sensitive to that variation.

    \rev{A second challenge in multi-finger manipulation is collecting high-quality data.
    In the real world, teleoperation for multi-finger hands \citep{qin2023anyteleop} and making use of human hand data \citep{xu2025dexumi, wi2026tactalign} are active areas of research.
    In contrast, while simulation allows for parallelizable data collection and easier variation of physical parameters \citep{qi2023hand}, teleoperation challenges are compounded and data generation policies may need to be specifically designed for different tasks.
    For more complex tasks, we may be limited in terms of data quality we can achieve, even in simulation.
    This motivates methods that can learn useful models from heterogeneous, or mixed quality, datasets.}

    In this work, we propose a method, \rev{Probabilistic Latent Unified world Modeling and parameter Estimation} (\namenospace), that learns a world model from datasets generated under diverse physical conditions.
    To enable richer reasoning over variations in dynamics, \rev{\name learns to estimate a belief over latent parameters, given observations and actions, and conditions world modeling on that belief.}
    Belief estimation allows us to deploy the same policy to situations with different dynamics, while \rev{planning allows for models trained on lower-quality datasets to still produce useful actions}.
    % Update the above

    Specifically, we train a flow matching \cite{lipmanflow} model that learns to both estimate parameters from task execution and uses that estimate to \rev{condition} a world model.
    We learn a unified latent representation of different parameters, such as friction coefficients, object shapes, and rewards, which are themselves functions of \rev{partially observable} variables.
    \rev{By training our representation end-to-end with the flow matching model, we learn representations that not only represent the parameter values, but are also useful conditioning for world model learning.}
    Through training our end-to-end model on data with varying parameter values, we enable parameter estimation through flow matching sampling for aligning our world model closer to the true dynamics at deployment without re-training.
    Finally, we construct a planner that auto-regressively \rev{samples} trajectories using our \rev{world} model and scores them by estimating rewards and the likelihoods of the transitions in the trajectory.

    Our contributions are (1) A unified flow-matching-based world model that estimates system parameters for use in conditioning world models. (2) A method to learn a unified latent representation over qualitatively different parameters end-to-end with our world model. (3) A planning method that uses world model rollouts and estimated likelihoods to optimize robot actions.

    We evaluate our method on multiple multi-finger manipulation tasks in simulation and on hardware.
    \rev{Across challenging simulated screwdriver turning, disk flicking, valve turning, and bucket lifting tasks, we significantly outperform state-of-the-art offline reinforcement learning and world model baselines.
    We also achieve zero-shot deployment of our simulation-trained model to a hardware precision screwdriver turning task.} 

%===============================================================================
\vspace{-.4cm}
\section{Related Work}
\vspace{-.4cm}
\label{sec:related_work}
\textbf{Generative Models for Manipulation:}
Generative models such as diffusion \cite{ddpm, song2020score} and flow matching \cite{lipmanflow} have been widely adopted for robotic manipulation.
These models approximate complex probability distributions over high-dimensional robot trajectories, making them attractive for planning and control.
They are often used to perform behavior cloning from expert demonstrations \cite{xu2025dexumi, chi2023diffusionpolicy, trajdiffuser, zhang2025flowpolicy}.
Specifically, in multi-finger manipulation, these models have also been applied to contact-rich planning \cite{kumar2025diffusing, kumar2025diffusion} and training large humanoid foundation models \cite{bjorck2025gr00t}.

Generative models have also been used to learn system dynamics through world models.
World models learn a dynamics function over observations, with prior work using that as an additional learning task to improve policy learning \cite{zhu2025unified, ye2026world, huang2025unified}.
Alternatively, they can be used for planning \citep{maes2026leworldmodel, kim2026cosmos} or reinforcement learning \citep{yin2026playworld}.
\rev{Unlike prior work, we} explicitly reason about variation in physical parameters that may lead to \rev{variation in dynamics}.
We evaluate against Unified World Model \citep{zhu2025unified}, which jointly \rev{learns a policy and world model} to improve behavior cloning.

\textbf{System Identification:}
System identification methods seek to estimate parameters of interest, including physical parameters.
Online reinforcement learning methods have integrated system identification components \citep{qi2023hand, kumar2021rma, hsieh2025learning} and have applied those methods to multi-finger manipulation.
However, these methods often require complex reward engineering and are trained online, whereas we explore learning from pre-generated offline datasets. 
In addition, these methods do not model uncertainty over parameter estimates and do not enable planning, unlike our method.
While there exists prior work that adapts ideas from classical Monte Carlo particle filtering \citep{dellaert1999monte} into data-driven methods \cite{jonschkowski2018differentiable}, including diffusion approaches \citep{wan2026diffpf}, these methods do not combine parameter estimation with downstream planning and focus on localization as opposed to physical parameter estimation.

\textbf{Offline Reinforcement Learning:}
Offline RL is concerned with learning policies from offline datasets, avoiding expensive environment interactions while being able to learn from heterogeneous data.
However, these methods \citep{fql_park2025, kostrikovoffline, kumar2020conservative, sinha2022s4rl} do not generally focus on the kind of joint system identification and policy learning we propose.
We compare against Flow Q-Learning (FQL) \cite{fql_park2025}, a SOTA offline reinforcement learning method that uses a flow-matching-based policy as well as Decision Diffuser \cite{ajayconditional}, a diffusion-based method that learns policies from heterogeneous data.

%===============================================================================

\begin{figure}[t]
    \centering
    \includegraphics[width=.99\linewidth]{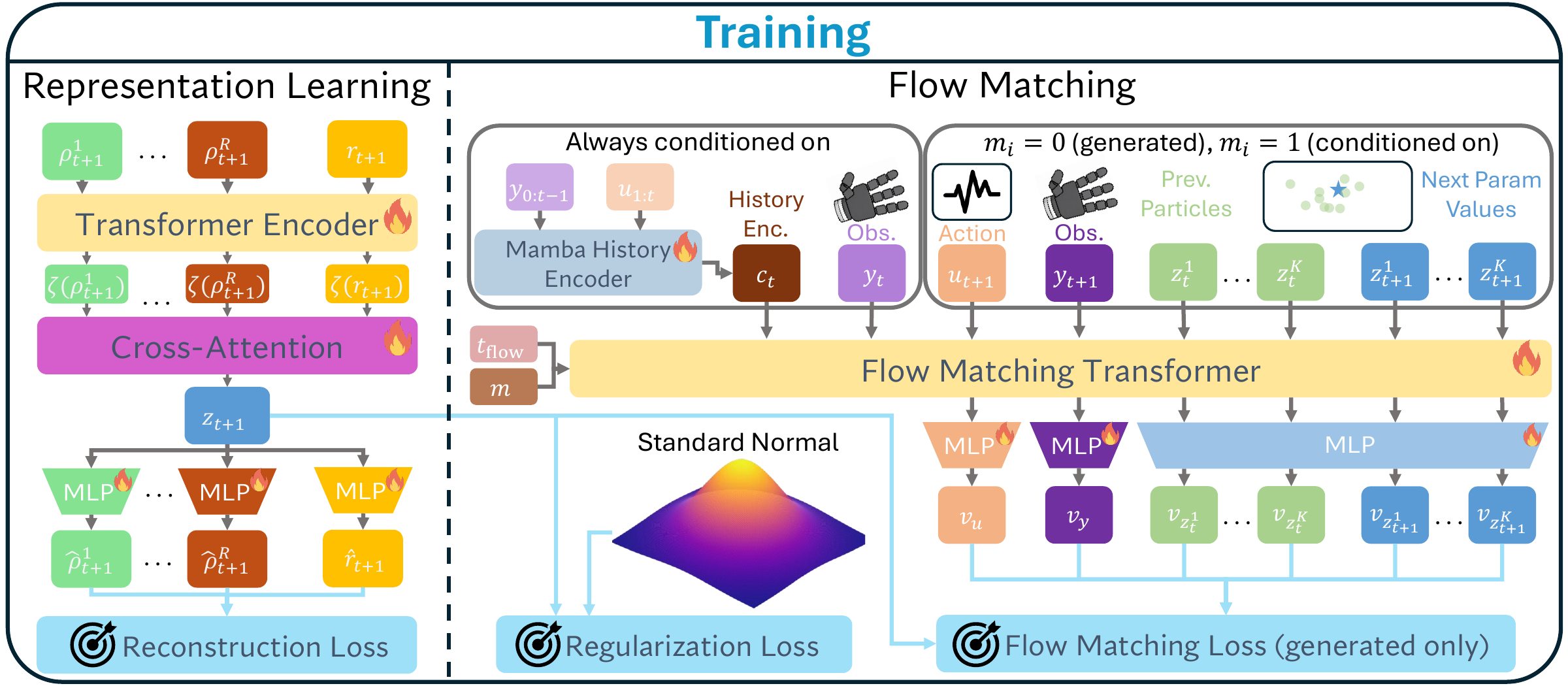}
    \caption{Our representation learning and flow matching (FM) architectures. 
    The transformer parameter encoder takes in a token for each parameter $\rho^i$ as well as the reward $r$ and uses cross-attention pooling of transformer features $\zeta$ to produce a latent embedding $z$.
    For the FM, we input all modeled variables in a transition and predict FM velocities $v$.
    We compute losses for $v_i$ where the conditioning mask $m_i = 0$.
    \rev{We use Rotary Position Embeddings \citep{su2024roformer} to distinguish timesteps.}
    }
    \label{fig:arch}
    \vspace{-.6cm}
\end{figure}

\vspace{-.3cm}
\section{Problem Statement}
\vspace{-.3cm}
\label{sec:problem_statement}
We consider the problem of performing manipulation in situations where there are $R$ relevant, potentially time-varying, system parameters $\rho_t = \{\rho_t^i\}_{i=1}^R$, which are uncertain.
These parameters could include friction coefficients, shape descriptors, or poses of manipulated objects.
These parameters \rev{induce uncertainty over} system dynamics and task rewards while not being directly observable.

We assume access to an offline dataset $\mathcal{D} = \{\tau_i\}_{i=1}^N$, where $\tau = \{y_{0:T}, r_{0:T}, \rho_{0:T}, u_{1:T}\}$ is a trajectory of length $T$ with observations $y_t$, rewards $r_t$, parameters of interest $\rho_t$, and actions $u_t$.
$y_t$ \rev{could include robot proprioception, noisy contact force measurements, noisy pose estimates, \rev{and/or} image observations}.
$u_{t+1}$ is the action taken to transition from timestep $t$ to $t+1$.
$\mathcal{D}$ can be generated in simulation, where true dense rewards can be computed and true parameter values are known and can be controlled.
% Given $\mathcal{D}$, our goal is to learn a model that can estimate unobservable system parameters during execution and use those estimates to inform planned trajectories.
\rev{Given $\mathcal{D}$, our goal is to learn a model that enables us to complete multi-finger manipulation tasks when parameters are uncertain.}

While access to a simulator would allow for applying methods like online reinforcement learning, learning a model-free policy to perform the types of precise tasks we evaluate on has been demonstrated to be challenging in prior work \cite{kumar2025diffusion, yang2025multi}, requiring significant simulation and reward engineering.
For more complex tasks, these methods may also require additional training data on the hardware system to train a useful policy \cite{hsieh2025learning}. 
In contrast, using an offline dataset allows us to use curated and informative data.
However, we do not restrict $\mathcal{D}$ to only have expert demonstrations.
We allow $\mathcal{D}$ to contain sub-optimal trajectories, which could include catastrophic failures such as dropping a screwdriver during a turning task.
Sub-optimal trajectories offer useful information about system dynamics, while reward labels enable us to use failure data for dynamics learning while avoiding problematic actions through planning.

%===============================================================================
% We reason primarily over transition tuples $(\rho_t, y_t, r_t, u_{t+1}, \rho_{t+1}, y_{t+1}, r_{t+1})$, where $u_{t+1}$ is the action taken to transition from timestep $t$ to $t+1$.
\vspace{-.3cm}
\section{Methods}
\vspace{-.3cm}
\label{sec:methods}

\begin{figure}[t]
    \centering
    \includegraphics[width=.89\linewidth]{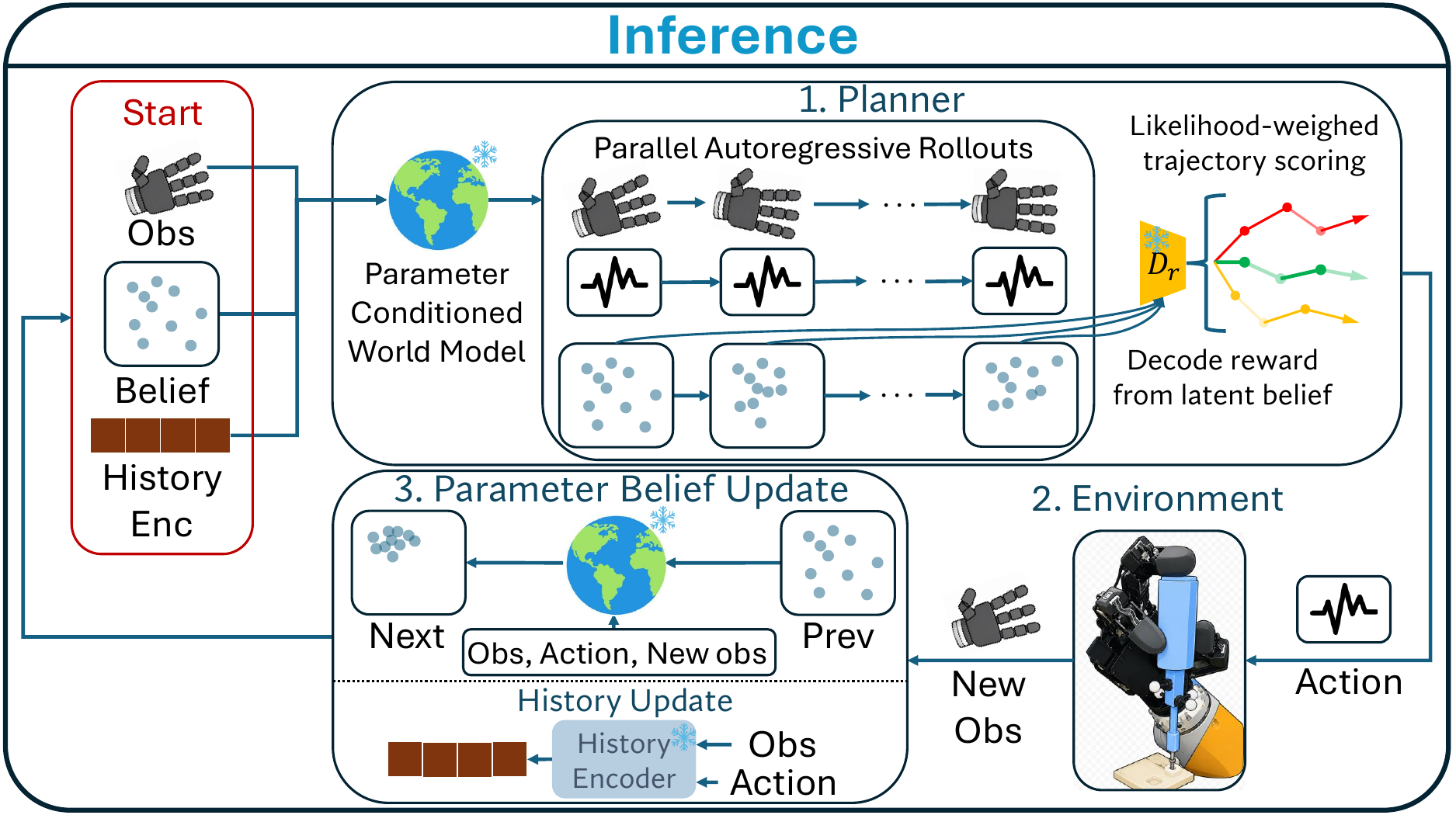}
    \caption{
    \rev{
        We initialize a latent belief by sampling several particles from a standard normal prior and initialize a history encoding as a zero vector.
        At each timestep, we use our world model to plan conditioned on the observation, belief, and history encoding.
        We score trajectories using their rewards decoded from the belief and their estimated likelihoods under the world model.
        We execute the first action from the selected plan. 
        After observing the resulting new observation, we update the latent belief with the world model.
        We then update the history encoding.
        This process repeats, allowing the world model to adapt without fine-tuning by updating belief over parameters.
        }
    }
    \label{fig:block_diagram}
    \vspace{-.6cm}
\end{figure}

We train a world model $M$ on the trajectory dataset $\mathcal{D}$, as shown in Fig.~\ref{fig:arch}.
Unlike prior world models, $M$ estimates a belief over a latent representation $z_t$ of hidden physical parameters and rewards, and uses that belief to condition its rollouts for planning. 
At deployment, $M$ maintains a belief $Z_t = \{\hat{z}_t^k\}_{k=1}^K$ over the latent encoding parameterized as a set of particles.
\rev{As shown in Fig.~\ref{fig:block_diagram}, we update $Z_t$ as we execute the task and use it to plan actions.
By estimating $Z_t$ online, we can adjust the dynamics of our world model \textit{without re-training}.}

Our method is designed to allow for end-to-end training of latent parameter representations and parameter-conditioned \rev{world modeling}.
We first define our model architecture, describe the training loop, then finally explain how we use our model for trajectory planning.

Our model architecture consists of an autoencoder for learning parameter representations \rev{(Sec.~\ref{sec:learning_param_rep}), a long-horizon history encoder (Sec.~\ref{sec:hist}), and a flow matching model (Sec.~\ref{sec:joint_fm})} that learns to both estimate latent parameters and model system dynamics conditioned on those parameters.

\vspace{-.2cm}
\subsection{Learning Parameter Representations}
\vspace{-.2cm}
\label{sec:learning_param_rep}

As we may want to model qualitatively different variables, for example physical parameters like friction coefficients along with shape representations or rewards, we propose learning a joint latent representation over these variables.
This also allows us to inform representation learning with flow matching gradients, shaping the latent representation to be most useful for learning system dynamics.
\rev{During training, we encode simulator-provided parameters and rewards into a latent target
$z_t = E(\tilde{\rho}_t)$, where $\tilde{\rho}_t = \rho_t \cup \{r_t\}$ and $E$ is a learned encoder.
We include $r_t$ in $\tilde{\rho}_t$ as rewards may be a function of variables, such as poses or velocities, that may not be directly observable.}

% \begin{wrapfigure}[12]{r}{0.4\linewidth}
%     \vspace{-4.5em}
%     \centering
%     \includegraphics[width=\linewidth]{corl_2026_template_submission/figs/Param_Encoder.png}
%     \captionsetup{font=footnotesize,skip=2pt}
%     \caption{The transformer parameter encoder, which takes in a separate token for each parameter as well as the reward and uses cross-attention pooling of transformer features $\zeta$ to produce a latent embedding $z$.}
%     \label{fig:param_enc}
%     \vspace{-1.2em}
% \end{wrapfigure}

We use a Wasserstein auto-encoder (WAE) \citep{tolstikhin2018wasserstein} to learn parameter representations.
The WAE imposes a standard normal prior that allows for more reasonable initialization of parameter belief from the standard normal.
For the encoder $E$, we use a transformer that takes in $\tilde{\rho}_t$ as input as shown in Fig.~\ref{fig:arch}.
We separately tokenize each $\tilde{\rho}^i_t$, and use an additive learned modality embedding to distinguish each element.
We use a cross-attention pooling layer to extract a single latent $z$ from the transformer output.
We train separate decoder MLPs $D_i$ for each parameter and the reward.

\vspace{-.3cm}
\subsection{History Modeling}
\vspace{-.3cm}
\label{sec:hist}
Standard particle-filtering-based methods use a Markov assumption, where actions and observations from previous timesteps are recursively compressed into the current belief.
This implicitly assumes the belief before the update retains all useful information from prior timesteps.
However, this may be challenging in situations with limited numbers of particles, noisy observations, or complex dynamics.
To retain useful historical context from previous trajectory timesteps, we use a Mamba-2 \citep{dao2024transformers} encoder $C$ to compress the trajectory history.
$C$ takes as input $u_{1:t}$ and $y_{0:t-1}$ and produces a history context vector $c_t$, with $c_0=0$.
We use $c_t$ as a summary of the full history excluding the most recent observation and action, which are directly passed into the flow matching model.

% \Needspace{18\baselineskip}

\begin{wraptable}{r}{.7\linewidth}
\vspace{-1.7em}
\centering
\footnotesize
\setlength{\tabcolsep}{0pt}
\renewcommand{\arraystretch}{0.92}
\caption{
Example conditional distributions learned by $M$.
}
\label{tab:distribution_modes}
\vspace{-0.9em}
\begin{tabularx}{\linewidth}{@{}
    >{\raggedright\arraybackslash}p{0.17\linewidth}
    >{\raggedright\arraybackslash}X
    >{\raggedright\arraybackslash}p{0.4\linewidth}
@{}}
\toprule
\textbf{Mode} & \textbf{Conditional Distribution} & \textbf{Purpose} \\
\midrule

\rev{Param. Estimation}
&
$p(Z_{t+1} \mid Z_t,\allowbreak y_t,\allowbreak u_{t+1},\allowbreak y_{t+1},\allowbreak c_t)$
&
Update latent belief \\

Rollout
&
$p(u_{t+1},\allowbreak y_{t+1},\allowbreak Z_{t+1}
\mid Z_t,\allowbreak y_t,\allowbreak c_t)$
&
Sample plans \\

Inverse dyn.
&
$p(u_{t+1} \mid Z_t,\allowbreak y_t,\allowbreak Z_{t+1},\allowbreak y_{t+1},\allowbreak c_t)$
&
Evaluate dynamics consistency of transitions \\

\bottomrule
\end{tabularx}
\vspace{-1.0em}
\end{wraptable}
\vspace{-.3cm}
\subsection{Joint Flow Matching}
\vspace{-.3cm}
\label{sec:joint_fm}

Flow matching learns a time-varying velocity field $v_\theta (x_t,t)$ that is integrated  to transform noise to representative samples $x$ from its training dataset.
Our model reasons over individual transitions and chains predictions autoregressively to allow for long-horizon rollouts.
% To enable $M$ to jointly learn , we pass all modeled variables to the transformer.
We train $M$ to jointly learn \rev{multiple conditional distributions from $\mathcal{D}$, including parameter estimation and rollout generation.
Table~\ref{tab:distribution_modes} includes example distributions learned by $M$.}

At trajectory time $t$, the model takes as input $y_t$,  $u_{t+1}$, $y_{t+1}, c_t$, $Z_t$, $Z_{t+1}$, and the flow matching timestep $t_{\mathrm{flow}}$.
\rev{Our flow matching model uses a transformer backbone, similar to DiT \citep{peebles2023scalable}, as shown in Fig.~\ref{fig:arch}.}
We specify the distribution we are targeting with an auxiliary mask input $m$, which is also passed into the model with $t_{\mathrm{flow}}$ through Adaptive LayerNorm \citep{peebles2023scalable}.
$m$ indicates which variables are conditioned on ($m_i=1$) and which variables are generated ($m_i=0$).
We pass in noisy inputs to the transformer when $m_i = 0$ and clean values when $m_i=1$.
This is similar to in-painting conditioning \cite{trajdiffuser} for flow matching or diffusion models, where part of a generated sample is provided as input and the remainder is generated conditioned on that input.
We pass the transformer features for the variables we are generating through an MLP to produce flow matching velocities $v$ used in flow matching loss calculation and ODE integration.
We only compute losses for generated variables ($m_i = 0$).

% \begin{wrapfigure}[11]{r}{0.35\linewidth}
%     \vspace{-1.2em}
%     \centering
%     \includegraphics[width=\linewidth]{corl_2026_template_submission/figs/corl_masks.png}
%     \captionsetup{font=footnotesize,skip=2pt}
%     \caption{A subset of distributions $M$ learns. Different masking patterns lead to different distributions.}
%     \label{fig:masks}
%     % \vspace{-1.2em}
% \end{wrapfigure}

\vspace{-.5cm}
\subsection{Training Loop}
\vspace{-.3cm}

We train \rev{the flow matching model $M$, the history encoder $C$, and the parameter encoder $E$ and decoders $D$} end-to-end on our dataset $\mathcal{D}$.
As our model takes in a set of previous particles as input, we perform filtering during the training loop using an exponential moving average (EMA) version of the model, emulating how the model is used at inference time.

Given a batch of training trajectories, we initialize $Z_0$ from $\mathcal{N}(0, I)$ and initialize $c_0$ as a zero vector.
We then iterate through the trajectory timesteps.
The full algorithm is in Appendix~\ref{app:training_loop}.
We sample random $m$ masks during training.
We always treat $ y_t, c_t$ as known, but randomly sample known/unknown booleans for $Z_t, Z_{t+1}, u_{t+1}, y_{t+1}$
\rev{from a Bernoulli distribution with $p=.5$ independently for each variable.}
For $Z_t$ and $Z_{t+1}$, we sample booleans at the level of the full particle set, not individual particles.

We train the WAE and the flow matching end-to-end, minimizing \rev{the loss in \eqref{eq:loss}, where $\hat{\tilde{\rho}}_t$ is the decoded true $z_t$ and $\beta$ is a loss weight:}

\vspace{-.5cm}
\begin{equation}
    \label{eq:loss}
    \mathcal{L} = ||\hat{\tilde{\rho}}_t - \tilde{\rho}_t||_2^2 + \beta \mathrm{MMD}(z_t) + \mathcal{L}_{\mathrm{FM}}
\end{equation} 
The first term is a reconstruction loss, the second is the MMD regularization described in \citet{tolstikhin2018wasserstein}, and the third term is the flow matching loss.
Minimizing the flow matching loss trains $M$ to estimate parameters and learn system dynamics conditioned on those parameters.
The regularization incentivizes the population of latent vectors to match a standard normal distribution, 
$\mathcal{L}_{FM}$ also guides latent vectors to be more useful for dynamics modeling.
\rev{Tying together the representation learning, parameter estimation, and world modeling within our end-to-end trained model augments our world model not only with the ability to model dynamics and generate actions, but also efficiently refine its dynamics predictions during task execution through parameter estimation.}

$\mathcal{L}_{FM}$ accumulates flow matching losses for different modeled variables from various conditional mask values.
This is potentially problematic for representation learning as allowing gradients for $Z$ estimation to backpropagate to the encoder could create a degeneracy where the flow matching is allowed to optimize its own target.
This would reduce the representational capacity in favor of creating an easier flow matching problem.
To address this, we do not update the WAE with gradients from the flow matching loss for $Z$.
This allows us to train the latent space in a way that better informs dynamics learning while retaining its representational capacity.

% \begin{figure*}[t]
%     \centering
%     \includegraphics[width=\linewidth]{figs/latent_flow_arch.png}
%     \caption{We show the architecture of the transformer used as the flow matching backbone, which includes modality embeddings $m$ to distinguish different input types as well as position encodings $p$ based on the trajectory timestep.}
%     \label{fig:arch}
% \end{figure*}

\vspace{-.5cm}
\subsection{Planning}
\vspace{-.3cm}
We use $M$ to plan actions.
Conditioned on a current $Z_t, y_t, c_t$, we generate planned trajectories of length $H$ by autoregressively sampling from the rollout distribution in Table~\ref{tab:distribution_modes}.
We update $c_t$ using generated actions and observations during the rollout using the trained Mamba-2 encoder $C$.
By sampling from the rollout distribution, we simultaneously model dynamics and parameter belief evolution over a trajectory.
\rev{As we iteratively update $Z_t$ during execution, we effectively modify the world model we use for planning \rev{as the estimated parameters change}.
This allows us to deploy a single model to situations with different dynamics and adapt to uncertainty during execution.}

\rev{We adopt a receding-horizon trajectory-sampling approach, where we sample a batch of $B$ trajectories from $M$, score the trajectories, and execute the first step of the highest scoring trajectory before replanning.
While we could exclusively score trajectories using the decoded rewards from the beliefs, that can lead to overly-optimistic plans that exploit errors in the learned world model.
To account for this, we incorporate the likelihoods of sampled transitions in our scoring.
}

We estimate rewards from the belief as $\hat{r}_t = \frac{1}{K}\sum_{k=1}^K D_r(z_t^k)$.
To approximate log-likelihoods $\ell_t$, we apply the method from \citet{zhou2023adaptiveonlinereplanningdiffusion}.
While \citep{zhou2023adaptiveonlinereplanningdiffusion} originally performs likelihood estimation for diffusion models using the diffusion evidence lower-bound loss, the flow matching loss can be used as a drop-in replacement.
Since we train $M$ to jointly model multiple probability distributions, we can choose the distribution under which we estimate likelihoods.
We find that using an inverse-dynamics likelihood estimation, specifically estimating $p(u_{t+1} | Z_t, y_t, Z_{t+1}, y_{t+1})$, empirically leads to the best performance.
A low inverse-dynamics likelihood indicates that a high-reward trajectory may have unrealistic dynamics and would therefore transfer poorly to actual execution.

We compute a score $s$ for each trajectory in the sampled batch as $s =\sum_{t=1}^H 
\gamma^t \hat{r}_t \rev{e^{\ell_t / \lambda}}$, where $\gamma$ is a discount factor and $\lambda$ is a tunable temperature.
We choose the trajectory with the highest $s$ score \rev{for execution}.
To avoid scoring issues when $r_t < 0$, we normalize rewards before training to [0, 1].

For the tasks we consider, the system may enter a state where $M$ struggles to model dynamics due to model error or the limited distribution covered by $\mathcal{D}$.
Continuing task execution at this point could lead to catastrophic failure.
We adopt prior work \citep{kumar2025diffusing} to address this, with more detail in Appendix~\ref{app:early_stop}

%===============================================================================

\vspace{-.5cm}
\section{Experimental Results}
\vspace{-.3cm}
\label{sec:result}
\rev{Our experiments are designed to investigate the utility of explicit parameter estimation, our end-to-end learning of the latent space and world model, and our planner.
To show \namenospace's utility, we focus on tasks that require contact-rich interaction with manipulated objects, often requiring precise interactions.
We evaluate on tasks where different physical properties of the system can lead to different outcomes, motivating the ability to reason over that variation. 
}

\rev{
Due to the sensitivity and complexity of contact-rich multi-finger manipulation, collecting high-quality datasets is a known challenge and an active research area
}\citep{xu2025dexumi, qin2023anyteleop, wi2026tactalign}.
\rev{
As such, being able to learn useful policies from datasets with mixed quality is an attractive property.
% than standard demonstration datasets \citep{khazatsky2024droid, liu2023libero}.
In our experiments, we show \namenospace's ability to outperform baselines on tasks with different levels of dataset quality.
}

We evaluate \name on simulated multi-finger screwdriver turning, valve turning, disk flicking, \rev{and bucket lifting (from \citet{bao2023dexart})} tasks in simulation and the screwdriver task in the real world.
For all tasks, we use an Allegro v4 hand with position PD control. \rev{We both generate data and run simulation experiments} in Isaac Lab \citep{mittal2025isaaclab} \rev{for all tasks except for bucket lifting, for which we use the SAPIEN simulator} \citep{xiang2020sapien}.
\rev{For the Isaac Lab tasks,} we use Isaac Lab contact normal force sensing to generate data in simulation, and use Xela sensors on the Allegro fingertips for hardware experiments.
We use a Vicon motion capture system for pose estimation for hardware experiments.

\vspace{-.3cm}
\subsection{Baselines and Ablations}
\vspace{-.3cm}

\rev{To evaluate our claims, we compare \name to multiple baselines and ablations.
For ablation details and results, please see Appendix~\ref{app:ablations}.}

\textbf{Unified World Model (UWM)} \citep{zhu2025unified}. UWM is a state-of-the-art (SOTA) behavior cloning (BC) method that unifies policy and world model learning, similar to \namenospace. \rev{To more closely match the training in the original paper, we filter out failures from our datasets only for UWM training.}
\textbf{Flow Q-Learning (FQL)} \citep{fql_park2025}. FQL is a SOTA offline reinforcement learning method that learns a one-step flow policy using offline Q-learning with behavior cloning regularization.
\textbf{Decision Diffuser (DD)} \citep{ajayconditional}. DD learns a return-conditioned diffusion model over observations and additionally learns an inverse dynamics model to predict actions.

While UWM, FQL, and DD learn generative policies (flow matching for FQL, diffusion for DD and UWM), they do not do explicit parameter estimation. 
In addition, none of the methods as published use inference-time scaling through planning as \name does, though DD generates longer-horizon trajectories and UWM learns a dynamics model that could be used for rollouts.
We train UWM, FQL, and DD with a fixed-length history of observations for implicit parameter reasoning.

% \rev{We also run several ablations of our design choices..}

\vspace{-.4cm}
\subsection{Tasks}
\vspace{-.4cm}
\rev{For all tasks, more details on task setup and dataset quality are in Appendix~\ref{app:task_setup}.}

\begin{figure}[t]
    \centering
    \includegraphics[width=.92\linewidth]{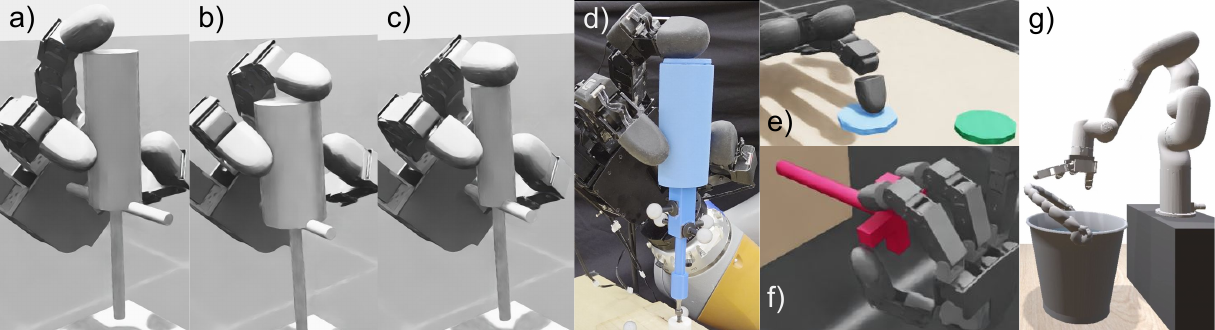}
    \caption{a-c) Simulated screwdriver task, showing random variation. d) Hardware screwdriver task. e) Disk flicking task, with randomly sampled goal. f) Valve turning task. g) DexArt bucket task.}
    \label{fig:sim_envs}
    \vspace{-.6cm}
\end{figure}

\textbf{Screwdriver Turning}. 
In this task, shown in Fig.~\ref{fig:sim_envs}a-d, the robot turns a precision screwdriver clockwise using the thumb, index, and middle fingers.
We vary screwdriver shape, finger-object friction, screw resistance friction, initial screwdriver pose, and initial grasp. We use shape descriptors and friction coefficients for $\rho$, and directly deploy the simulation-trained model on hardware (Fig.~\ref{fig:sim_envs}d) with no fine-tuning or hardware-specific planner tuning.

% \begin{wrapfigure}[17]{r}{0.25\linewidth}
%     \vspace{-1.2em}
%     \centering
%     \includegraphics[width=\linewidth]{corl_2026_template_submission/figs/hardware_screwdriver_only.pdf}
%     \captionsetup{font=footnotesize,skip=2pt}
%     \caption{Hardware screwdriver turning task}
%     \label{fig:hw_screw}
%     % \vspace{-1.2em}
% \end{wrapfigure}

% We directly deploy the simulation-trained model on hardware, as shown in Fig.~\ref{fig:sim_envs}d, with no fine-tuning on hardware data.
% In addition, we directly reuse planning hyperparameters tuned in simulation, without specific tuning on the hardware system.

\textbf{Disk Flicking}. In this task, shown in Fig.~\ref{fig:sim_envs}e, the robot flicks a disk with its index finger to a random  goal location, with random friction coefficient between the finger and the disk and between the disk and the table.
\rev{We use these friction coefficients for $\rho$.}
% \begin{wrapfigure}[17]{r}{0.4\linewidth}
%     \vspace{-1em}
%     \centering
%     \includegraphics[width=\linewidth]{corl_2026_template_submission/figs/sim_env_disk_valve.pdf}
%     \captionsetup{font=footnotesize,skip=2pt}
%     \caption{a) Disk flicking task, with goal colored green. b) Valve turning task}
%     \label{fig:sim_env}
%     % \vspace{-1.2em}
% \end{wrapfigure}
As we lose control over the disk after the flicking motion, high-error dynamics modeling can lead to missing the goal.

\textbf{Valve Turning}. In this task, shown in Fig.~\ref{fig:sim_envs}f, the hand turns a valve using its thumb, index, and middle fingers, where we randomly set the friction between the fingers and the valve and the friction of the valve joint and use those frictions for $\rho$.
We also randomly initialize the hand configuration and the valve angle.
We additionally demonstrate \name on a hardware valve, with more details in Appendix~\ref{app:valve_turning}.

\rev{
\textbf{DexArt Bucket}. In this task, shown in Fig.~\ref{fig:sim_envs}g, a 6 DOF arm with mounted Allegro hand grasps the articulated handle of a bucket and lifts it.
The bucket used in each episode is randomized from a set of 11 distinct shapes.
We use the shape descriptor for $\rho$ and use the original DexArt reward.
}

\vspace{-.5cm}
\subsection{Results}
\vspace{-.3cm}
We show results for \name compared to baselines in Table~\ref{tab:results}.
We run 150 trials for the simulation tasks and 30 trials for the hardware task.
For screwdriver turning, we consider Validity Rate, meaning avoiding dropping the screwdriver or unmating it from the screw, to be the primary metric with lower goal distance being secondary.
In an overarching system which sequences multiple turning primitives, having a higher validity rate should lead to further turning over time.

In the simulated screwdriver experiments, \name achieves significantly higher validity than baselines ($p < .05$) with a McNemar test \citep{mcnemar1947note}, \rev{which compares categorical outcomes like validity/invalidity}.
This shows \namenospace's ability to generalize across diverse screwdriver shapes and physical parameters in simulation.
We achieve a 10.6 percentage point improvement in validity over the best baseline (UWM), while also achieving a 31\% reduction in goal distance.
% In addition, \name achieves a slightly higher validity rate and a significantly lower goal distance ($p<.05$) using a Welch's t-test \citep{welch1947generalization} than applying our data generation method, showing \name can outperform heterogeneous data.

On hardware, \name achieves a higher validity rate than baselines, with the 2 failures out of 30 trials resulting from the hand unmating the screwdriver from the screw as opposed to dropping it.
While FQL does achieve a lower distance to goal on its one valid trial, the validity rate is much lower.
This is partially due to it failing to settle after turning.
In contrast, \name can reason about the quality of its planning through OOD detection to avoid dropping.
We achieve a 6.6\% higher validity rate and a 10\% lower goal distance than the UWM baseline.
We observe that \name achieves a lower goal distance earlier in execution than UWM, \rev{discussed further in Appendix~\ref{app:hw_results}.}

For the disk and valve tasks, \name significantly outperforms all baselines \rev{using a Welch's t-test \citep{welch1947generalization}}.
We achieve a 54\% lower distance to goal than UWM for the valve and a 20\% lower distance to goal for the disk flicking.
\rev{We find the baselines struggle to generalize across the varying conditions we evaluate on, especially given the forceful, dynamic motion needed for the disk flicking task.}

% \begin{wraptable}{t}{0.4\columnwidth}
% \vspace{-.5cm}
% \footnotesize
% \centering
% \setlength{\tabcolsep}{1.7pt}
% \renewcommand{\arraystretch}{0.95}

% \begin{tabular}{@{}p{1.3cm}p{2.7cm}c@{}}
% \Xhline{2pt}
% & \centering Method
% & \makecell{Goal\\Distance\\(cm) $\downarrow$}\\
% \hline
% \multirow{4}{*}{\parbox{1.3cm}{\centering Sim. Disk\\Flicking}}
% & \name & $\mathbf{6.5 \pm .7}$\\
% & FQL \cite{fql_park2025} & $9.8 \pm 1.9$\\
% & DD \cite{ajayconditional} & $8.6 \pm 1.4$\\
% & UWM \cite{zhu2025unified} & $8.1 \pm .7$\\
% \hline
% \end{tabular}

% \caption{Disk flicking results. 95\% confidence intervals for goal distance.}
% \label{tab:results_disk}
% \vspace{-0.5cm}
% \end{wraptable}

% Our results show \name can outperform a heterogeneous dataset for the screwdriver turning task, though does not match the performance of the higher quality datasets for the disk and valve task.
% We note that \name still provides the best performance when compared to the baselines.
\begin{table*}[tbp]
\centering
\footnotesize
\setlength{\tabcolsep}{2.2pt}
\renewcommand{\arraystretch}{0.95}

\begin{tabular}{@{}>{\raggedright\arraybackslash}p{2.05cm}ccccccc@{}}
\Xhline{2pt}
\multirow{2}{*}{\centering Method}
&
\multicolumn{2}{c}{\makecell{Sim.\\Screwdriver}}
&
\multicolumn{2}{c}{\makecell{Hardware\\Screwdriver}}
&
\makecell{Sim.\\Valve}
&
\makecell{Sim. Disk\\Flicking}
&
\makecell{DexArt\\Bucket}
\\
\cmidrule(lr){2-3}
\cmidrule(lr){4-5}
\cmidrule(lr){6-6}
\cmidrule(lr){7-7}
\cmidrule(l){8-8}
&
\makecell{Validity $\uparrow$}
&
\makecell{Goal Dist. $\downarrow$}
&
\makecell{Validity $\uparrow$}
&
\makecell{Goal Dist. $\downarrow$}
&
\makecell{Goal Dist. $\downarrow$}
&
\makecell{Goal Dist. $\downarrow$}
&
\makecell{Success $\uparrow$}
\\
\hline

\namenospace
& $\mathbf{81.3\%}$ & $\mathbf{.88 \pm .09}$
& $\mathbf{93.3\%}$ & $\mathbf{.77 \pm .18}$
& $\mathbf{.59 \pm .09}$
& $\mathbf{6.5 \pm .7}$
& $\mathbf{71.3\%}$
\\

UWM \cite{zhu2025unified}
& $70.7\%$ & $1.27 \pm .09$
& $86.7\%$ & $.86 \pm .23$
& $1.29 \pm .05$
& $8.1 \pm .7$
& $0\%$
\\

FQL \cite{fql_park2025}
& $21.3\%$ & $.48 \pm .13$
& $3.3\%$ & $.07$
& $1.52 \pm .07$
& $9.8 \pm 1.9$
& $5.3\%$
\\

DD \cite{ajayconditional}
& $19.3\%$ & $.46 \pm .11$
& $48.3\%$ & $.95 \pm .12$
& $1.34 \pm .04$
& $8.6 \pm 1.4$
& $0\%$
\\

\Xhline{2pt}
\end{tabular}

\caption{Task performance results.
For the screwdriver tasks, goal distances are bolded for the method with the best validity rate.
Goal distance is reported in radians for screwdriver and valve, and centimeters for disk flicking.
All goal-distance intervals are 95\% confidence intervals.}
\label{tab:results}
\vspace{-.6cm}
\end{table*}

% \begin{wraptable}{t}{0.4\columnwidth}
% \vspace{-.5cm}
% \footnotesize
% \centering
% \setlength{\tabcolsep}{1.7pt}
% \renewcommand{\arraystretch}{0.95}

% \begin{tabular}{@{}p{1.3cm}p{2.7cm}c@{}}
% \Xhline{2pt}
% & \centering Method
% & \makecell{Success $\uparrow$}\\
% \hline
% \multirow{4}{*}{\parbox{1.3cm}{\centering DexArt\\Bucket}}
% & \name & $\mathbf{71.3\%}$\\
% & UWM \cite{zhu2025unified} & $0\%$\\
% & FQL \cite{fql_park2025} & $5.3\%$\\
% & DD \cite{ajayconditional} & $0\%$\\
% \hline
% \end{tabular}

% \caption{DexArt bucket results.}
% \label{tab:results_dexart}
% \vspace{-0.5cm}
% \end{wraptable}

\rev{
\name is the only method that consistently succeeds at the DexArt bucket task.
Our results suggest that our method can reason about partially observable state variables, including position, velocity, and orientation in addition to physical parameters.
Because rewards depend on these state variables, reward estimation that is sufficient to support downstream planning provides evidence of this capability.
For example, the DexArt reward is a function of the bucket position, the handle articulation angle, and the contact interaction between the hand and the bucket.
}

% Something about how \name qualitatively has better looking trajectories. Need a figure showing that

% \name significantly outperforms the DD baseline and most ablations on the disk flicking task, while technically not having a statistically significantly lower distance than FQL due to the high variance of FQL performance.

% We find that \name is the only one which is in the group of statistically best performing methods across all tasks.

%===============================================================================

\vspace{-.5cm}
\section{Limitations}
\vspace{-.3cm}
\label{sec:limitations}

	The main limitation of our method is the need for ground-truth parameter values for training.
    While these values are accessible in simulation, they may not be known in the real world.
    We could potentially incorporate real robot or human data by fine-tuning a model pre-trained on simulation data.
    For example, the pre-trained model could  label a real dataset with parameter values and then fine-tune the model with the actions and observations from the real-world dataset.
    This would align the pre-trained prior which can represent many different physical variations with the subset needed at inference time without requiring ground truth labels.
    Future work could explore this fine-tuning.

    \rev{An additional limitation is the quadratic scaling of memory usage by transformers as the number of input tokens increases.
    This limits the number of particles we can use for estimation.
    However, this could be addressed with alternate architectures \cite{sun2023retentive, de2024griffin} that provide sub-quadratic memory scaling.}

%===============================================================================

\vspace{-.3cm}
\section{Conclusion}
\vspace{-.3cm}
\label{sec:conclusion}

We presented a method for \rev{multi-finger} manipulation under uncertainty over hidden \rev{physical} parameters \rev{like object shapes and finger/object friction coefficients}.
Our method learns a latent representation from simulator-provided parameters and rewards, maintains a particle-approximated belief over true parameter values at deployment, and uses a single \rev{unified} flow matching model for both belief updates and \rev{belief-conditioned world modeling}.
By combining reward prediction with likelihood-based trajectory scoring, \rev{we construct a planner that uses rollouts from the world model to generate actions.}
Across simulated screwdriver turning, valve turning, disk flicking, and bucket lifting tasks, as well as zero-shot transfer to real-world screwdriver turning, our method improves over offline reinforcement learning, diffusion-based planning, and world modeling baselines.
\rev{Our results suggest, for the precise and sensitive tasks we consider, that estimating and conditioning on explicit representations of system parameters leads to better performance than methods that try to learn policies agnostic to that variation or try to model it implicitly.}

%===============================================================================

\clearpage
% The acknowledgments are automatically included only in the final and preprint versions of the paper.
% \acknowledgments{If a paper is accepted, the final camera-ready version will (and probably should) include acknowledgments. All acknowledgments go at the end of the paper, including thanks to reviewers who gave useful comments, to colleagues who contributed to the ideas, and to funding agencies and corporate sponsors that provided financial support.}

%===============================================================================

\bibliography{example}  % .bib

\appendix
\appendixpage

\section{Attention Masking}
\rev{By default, all tokens in the flow matching transformer attend to each other.
However, when updating our belief, allowing generated particles to attend to each other can lead to all generated particles collapsing to the same value.
To ensure that our belief captures uncertainty over parameter values, we mask the transformer attention to preserve independent sampling of each $\hat{z}_t^k$. 
We do not allow any of the generated $\hat{z}_t^k$ particles to attend to each other.
However, this is not sufficient to prevent communication between generated particles.
To further control the interaction between different tokens in the transformer, we also do not allow any known tokens to attend to unknown tokens.
}

\section{Training Loop}
\label{app:training_loop}

\begin{algorithm}[H]
\caption{Training Loop}
\label{alg:training_loop}
\begin{algorithmic}[1]
\Require Dataset $\mathcal{D}$, Flow Matching model $M$, EMA model $M_{\mathrm{EMA}}$, Parameter encoder $E$, Parameter decoders $D$, Mamba-2 history encoder $C$

\State $Z_0 \sim \mathcal{N}(0,I)$
\State $c_0 \gets 0$

\For{$0 \leq t \leq H-1$}
    \State $m \gets$ randomly sample training mask
    \State $z_{t+1} \gets E(\tilde{\rho}_{t+1})$
    \State $\hat{\tilde{\rho}}_{t+1} \gets D(z_{t+1})$
    \State $\mathcal{L}_{\mathrm{FM}} \gets \mathtt{FM\_loss}(m, Z_t, z_{t+1}, y_t, y_{t+1}, u_{t+1})$
    \State $\mathcal{L} \gets ||\hat{\tilde{\rho}}_{t+1} - \tilde{\rho}_{t+1}||_2^2 + \beta \cdot MMD(z_t) + \mathcal{L}_{\mathrm{FM}}$
    \State Update $M, E, D$ weights
    \State Update $M_{\mathrm{EMA}}$
    \State $Z_{t+1} \sim M_{\mathrm{EMA}}(Z_{t+1} \mid Z_t, y_t, u_{t+1}, y_{t+1}, u_t)$
    \State $c_{t+1} \gets C(u_{1:t}, y_{0:t-1})$
\EndFor

\end{algorithmic}
\end{algorithm}

\section{Out-of-Distribution Detection}
\label{app:early_stop}
\citet{kumar2025diffusing} propose an out-of-distribution (OOD) stopping criterion for task execution based on the likelihoods of generated plans. 
We adopt a similar approach in our planner.
We define the log-likelihood of a generated trajectory as the sum of each transition's log-likelihood, again using the inverse-dynamics likelihood.
If the average trajectory log-likelihood is less than some threshold $\ell_{\mathrm{min}}$, we stop executing.
The OOD thresholding provides a heuristic for identifying states from which the model cannot confidently model dynamics and its plans therefore cannot be trusted.
$\ell_{\mathrm{min}}$ can be tuned by running trials without early stopping and post-hoc analyzing the trajectories to find the optimal value.

\section{Task Setup}
\label{app:task_setup}

As our original simulation dataset is noise-free, we add noise during training and simulation evaluation to $y_t$ for the screwdriver, disk flicking, and valve turning tasks to more closely match real-world evaluation.
We add noise to \rev{object} pose observations as well as sensed contact forces.
We do not add noise during hardware evaluation.

\subsection{Screwdriver Turning}
The base of the screwdriver is fixed to the table but is free to rotate in 3 dimensions, simulating driving a screw in a slot.
We parameterize the screwdriver orientation $o$ with roll, pitch, and yaw Euler angles.
The goal is to turn the screwdriver $\frac{\pi}{2}$ rad. clockwise while keeping the roll and pitch angles 0 (keeping the screwdriver upright), though this may not always be feasible given the initialization.
\rev{The goal $g$ is therefore $[0, 0, o^{\mathrm{yaw}}_0 - \frac{\pi}{2}]$.
Our reward is the negative SO(3) angle distance from the current screwdriver orientation to the goal orientation:
\begin{equation}
    r(R_o, R_g) = -\cos^{-1}\left( \frac{\operatorname{tr}(R_o^\top R_g)-1}{2} \right)
\end{equation}
where $R_o, R_g$ are the rotation matrix representations of $o, g$ respectively.}
\rev{As \name estimates the reward, which is a function of the pose, we do not find it necessary to explicitly estimate the screwdriver pose.}

We vary shape by randomly sampling the height and diameter of the screwdriver body.
We pre-train a 32 dimensional Neural Descriptor Field (NDF) \citep{simeonov2022neural} shape representation to model the different screwdriver shapes.

To simulate the increase in resistance that occurs when a screw is tightened, we increase the yaw joint friction as the screwdriver turns, linearly interpolating from the initial sampled value to the maximum friction value over the course of the 90-degree turn.

We generate a training dataset of 8059 trajectories, each with 24 actions.
To generate training data, we use \citet{yang2025multi}, providing the trajectory optimization the true shape, pose, and physical parameters.
\rev{This leads to heterogeneous trajectory quality in $\mathcal{D}$, with 11.9\% trajectories dropping the screwdriver and non-dropped trajectories achieving average distance to goal of 1.25 rad.}

We run evaluations for all methods for a maximum of 24 steps, with \name terminating if it goes OOD.

\subsection{Disk Flicking}
We use a weighted sum of goal distance and velocity as the reward: $r_t = -||p_t - g||_2 - .1 \cdot ||v_t||_2$, where $p_t$ is the disk position, $g$ is the goal position, and $v_t$ is the disk velocity.

We train an oracle goal-conditioned reinforcement learning policy with PPO \citep{schulman2017proximal} for data collection.
We collect 8100 training rollouts from the oracle policy, with 32 timesteps for training data and evaluation.
Once we collect rollouts from the oracle policy, we relabel the goal for each trajectory as its end state.
\rev{With this relabeling, we construct a dataset closer to standard demonstration datasets where training trajectories are guaranteed to reach the goal.}
We pass in $g$ as an additional \rev{conditioning token} input to $M$, include it as part of the observation for FQL and UWM, and use goal-conditioning instead of return conditioning for \rev{the Decision Diffuser baseline}.

\subsection{Valve Turning}
\label{app:valve_turning}
The goal is to turn the valve to $-\frac{\pi}{2}$ rad.
The reward is the yaw distance to the goal.
Similarly to the screwdriver, we increase the joint friction as the valve turns.
We generate 7670 training trajectories with 32 timesteps.

We use an oracle Model Predictive Path Integral Control (MPPI) \citep{williams2017information} policy, informed with the true Isaac simulator dynamics for data generation.
\rev{The average distance to goal for dataset trajectories is .09 rad, leading to an overall high-quality dataset that still contains slight variation in task performance.}

\begin{wrapfigure}[14]{r}{0.5\textwidth}
    \vspace{-2.6em}
    \centering
    \includegraphics[width=\linewidth]{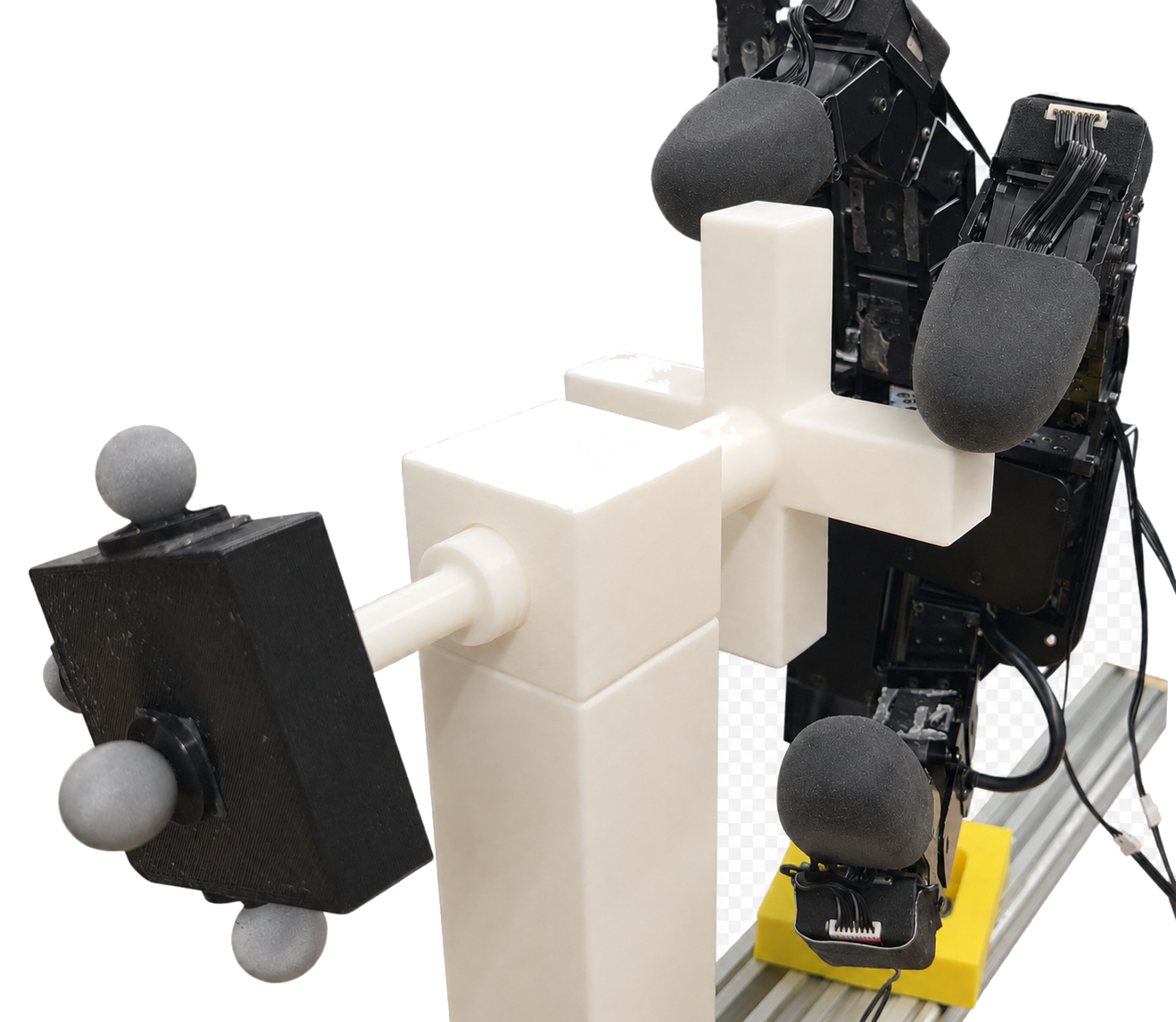}
    \captionsetup{font=footnotesize,skip=2pt}
    \caption{Hardware valve task}
    \label{fig:hw_valve}
    \vspace{-1.2em}
\end{wrapfigure}

On our website, we provide multiple successful demonstrations of \name on a hardware version of the valve turning task.
Like the screwdriver, we use a Vicon mocap system for valve angle estimation, as shown in Fig.~\ref{fig:hw_valve}.
Similar to the simulation version, we randomly initialize robot joint angles and the valve angle.
As in simulation, the task objective is to turn the valve to a $\frac{\pi}{2}$ radian orientation.

\subsection{DexArt Bucket}
The DexArt reward is a function of the bucket's pose, the articulation angle of the handle, and the contact interaction between the hand and bucket.
The DexArt task evaluates our method on a higher DoF system, as it involves 16 DoF across all 4 fingers and the 6 DoF of the arm. 
We use image observations, making use of frozen, pre-trained \citep{simeoni2025dinov3} features.
Specifically, we concatenate the \texttt{[CLS]} token output with the mean of all patches,
along with proprioception consisting of 22 DoF joint positions as well as linear velocity, angular velocity, position, and pose of the palm to construct a single feature vector for $y_t$.

Similar to the screwdriver, we pre-train an NDF representation on the set of buckets and estimate the shape embedding with our method. Specifically, we use the "seen" set of buckets from \citet{bao2023dexart}.
We generate 900 trajectories using the pre-trained policy from the original DexArt paper.
$\mathcal{D}$ has a success rate of 62.3\%, meaning there is significant heterogeneity.

\section{Ablations}
\label{app:ablations}
We run multiple ablations of our design choices.
\textbf{Single Particle} uses $K=1$ for parameter estimation, reducing the multi-particle belief to a point estimate.
\textbf{No Belief} adopts a more traditional approach for joint future action and observation generation.
No Belief only estimates scalar rewards using a single particle without using a latent space and uses those rewards and generated rollouts for planning.
\textbf{Detached} does not inform latent space learning using flow matching gradients.
\textbf{No Early Stop} does not use likelihood thresholding for stopping execution to evaluate the efficacy of early stopping and to provide a fairer comparison to the baselines, which do not have early stopping functionality.
\textbf{No Likelihood} ablates the use of likelihood entirely from the planner, meaning we do not weigh rewards by likelihood or do early stopping.

We show that our early stopping leads to minor improvement in performance for the simulated screwdriver, though \name still has higher validity rates than the baselines without it.
For the disk flicking, valve turning, and DexArt, as there is less of a validity concern, we can set an arbitrarily low stopping threshold and therefore observe no difference from early stopping.

While \name significantly outperforms most ablations for the valve, we observe less significant difference to the ablations for the disk flicking task.
This may be due to the finger spending less time in contact with the disk, thereby collecting less information for use in parameter estimation.
Future work could explore integrating active data collection at deployment, potentially by modifying the planner, to collect more informative transitions to improve performance.
We also observe that belief estimation is most helpful for the screwdriver task, which is the most complex task due to the combination of friction and shape variation along with the potential for catastrophic failure through dropping.

When considering all tasks, \name outperforms the ablations.

\begin{table*}[tbp]
\centering
\footnotesize
\setlength{\tabcolsep}{2.6pt}
\renewcommand{\arraystretch}{0.95}

% Put this in the preamble if you use it elsewhere:
\newcolumntype{C}[1]{>{\centering\arraybackslash}p{#1}}

\begin{tabular}{@{}>{\raggedright\arraybackslash}p{2.15cm}C{1.55cm}C{1.55cm}C{1.55cm}C{1.55cm}C{1.35cm}@{}}
\Xhline{2pt}
\multirow{2}{*}{Method}
&
\multicolumn{2}{c}{\makecell{Sim.\\Screwdriver}}
&
\makecell{Sim.\\Valve}
&
\makecell{Sim. Disk\\Flicking}
&
\makecell{DexArt\\Bucket}
\\
\cmidrule(lr){2-3}
\cmidrule(lr){4-4}
\cmidrule(lr){5-5}
\cmidrule(l){6-6}
&
\makecell{Validity\\Rate $\uparrow$}
&
\makecell{Goal\\Dist. $\downarrow$}
&
\makecell{Goal\\Dist. $\downarrow$}
&
\makecell{Goal\\Dist. $\downarrow$}
&
\makecell{Success\\Rate $\uparrow$}
\\
\hline

\namenospace
& $\mathbf{81.3\%}$ & $\mathbf{.88 \pm .09}$
& $\mathbf{.59 \pm .09}$
& $\mathbf{6.5 \pm .7}$
& $\mathbf{71.3\%}$
\\

No Early Stop
& $76.0\%$ & $.95 \pm .05$
& $\mathbf{.59 \pm .09}$ 
& $\mathbf{6.5 \pm .7}$
& $\mathbf{71.3\%}$
\\

Single Particle
& $\mathbf{82.7\%}$ & $1.36 \pm .08$
& $.72 \pm .09$
& $7.7 \pm 1.2$
& $63.3\%$
\\

No Belief
& $69.3\%$ & $.63 \pm .08$
& $\mathbf{.52 \pm .08}$
& $\mathbf{6.5 \pm 1.3}$
& $\mathbf{67.3\%}$
\\

Detached
& $\mathbf{85.3\%}$ & $\mathbf{.80 \pm .08}$
& $.69 \pm .09$
& $\mathbf{6.7 \pm .8}$
& $55.3\%$
\\

No Likelihood
& $46.7\%$ & $.48 \pm .08$
& $.68 \pm .09$
& $\mathbf{6.8 \pm .7}$
& $\mathbf{68.7\%}$
\\

\Xhline{2pt}
\end{tabular}

\caption{Ablation results. Bolded values do not statistically significantly underperform the absolute best value. Goal distances are in radians for screwdriver/valve and centimeters for disk flicking. Intervals are 95\% confidence intervals.}
\label{tab:ablations}
\vspace{-0.6cm}
\end{table*}

\section{Further Hardware Analysis}
\label{app:hw_results}

\begin{wrapfigure}[14]{r}{0.5\textwidth}
    \vspace{-2.6em}
    \centering
    \includegraphics[width=\linewidth]{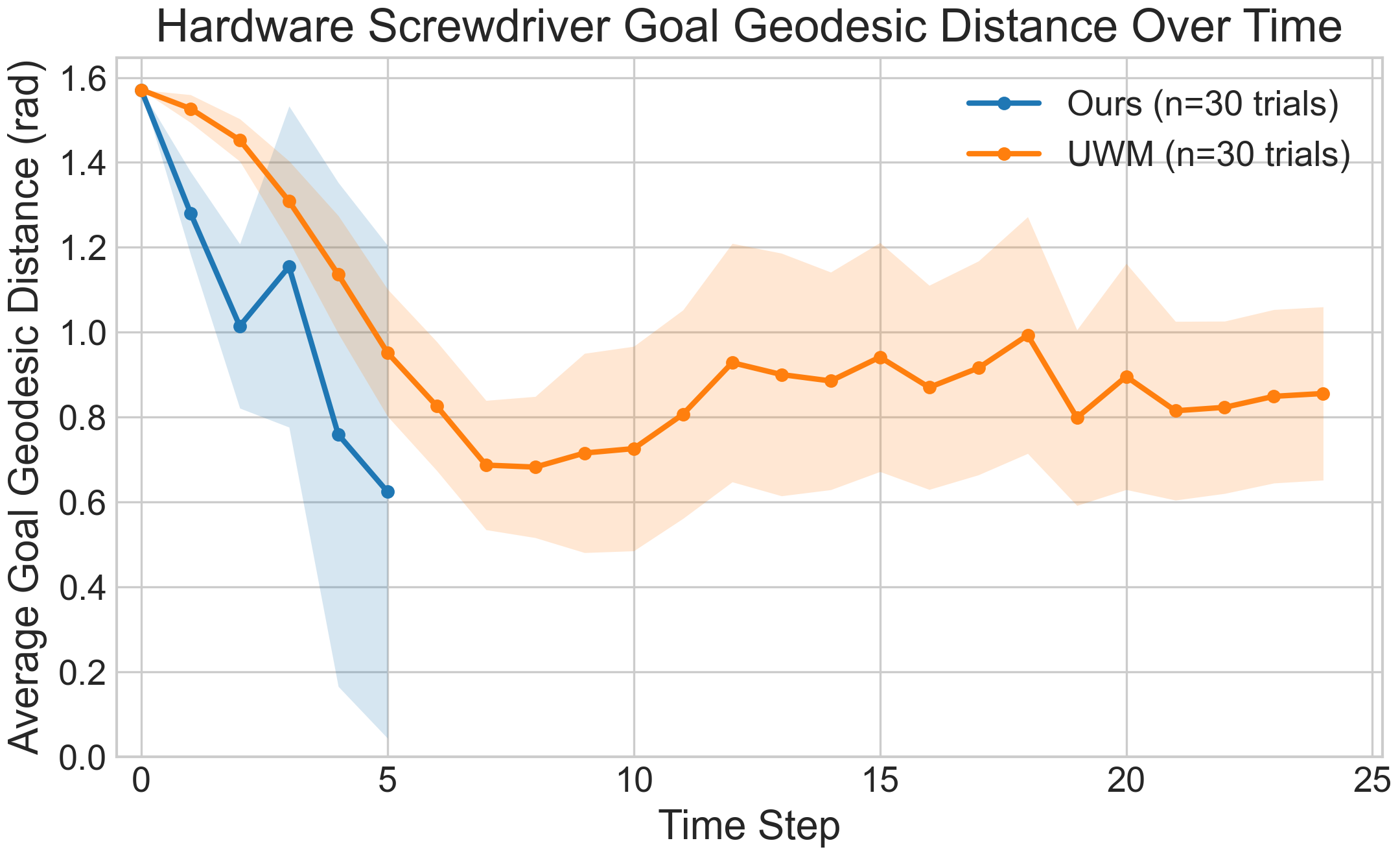}
    \captionsetup{font=footnotesize,skip=2pt}
    \caption{Distance to goal over time for the hardware screwdriver turning task, with shaded 95\% CI. \name stops executing after achieving a low distance to goal, avoiding an increase later in execution.}
    \label{fig:hw_screwdriver_line}
    \vspace{-1.2em}
\end{wrapfigure}

Fig.~\ref{fig:hw_screwdriver_line} shows the benefit of our early stopping on hardware, where we achieve a lower distance to goal earlier than UWM rollouts and are able to stop the policy at a lower distance to goal.
In contrast, UWM exhibits a slight increase in distance later on in the trajectories.
\name only requires on average 2.6 actions per trial.

\section{Timings}
Our planning times are .26 s per action for the screwdriver, .37 s for the valve, .18 s for the disk flicking, and \rev{.68} s for the bucket lifting.
\rev{Particle update times are .03 s for the screwdriver, .05 s for the valve, .03 s for the disk flicking, and .03 s for the bucket lifting}.
Model training takes approximately 8 hours on 2 RTX 5090 GPUs.

% \section{Hyperparameters}
\end{document}